\documentclass[pmlr,twocolumn,10pt]{jmlr} % W&CP article

\usepackage{listings}
\usepackage{xcolor}
\usepackage{tcolorbox}
\usepackage{siunitx}
\usepackage{booktabs}
\usepackage{multirow}
\usepackage{comment}
\tcbuselibrary{breakable}
\usepackage[switch]{lineno}

\theorembodyfont{\upshape}
\theoremheaderfont{\scshape}
\theorempostheader{:}
\theoremsep{\newline}

\newcommand{\method}{\texttt{POET }}

\jmlrvolume{LEAVE UNSET}
\jmlryear{2026}
\jmlrsubmitted{LEAVE UNSET}
\jmlrpublished{LEAVE UNSET}
\title[\texttt{POET}]{\texttt{POET}: Protocol Optimization via Eligibility Tuning}

\author{%
\Name{Trisha Das} \Email{trishad2@illinois.edu}\\
\addr University of Illinois Urbana-Champaign
\AND
\Name{Katherine Kero} \Email{katherine.kero@optum.com}\\
\addr Optum
\AND
\Name{Dorinda Schumann} \Email{dori\_schumann@optum.com}\\
\addr Optum
\AND
\Name{Tracy Ohrt} \Email{tracy.ohrt@optum.com}\\
\addr Optum
\AND
\Name{Sanjit Singh Batra}\nametag{$^*$} \Email{sanjit.batra@optum.com}\\
\addr Optum AI
\AND
\Name{Gregory D Lyng}\nametag{$^*$} \Email{Gregory.Lyng@optum.com}\\
\addr Optum AI
\AND
\Name{Robert E. Tillman}\nametag{$^*$} \Email{rob.tillman@optum.com}\\
\addr Optum AI
\vfill
\centerline{$^*$Corresponding authors}
}
\begin{document}

\maketitle

\begin{abstract}
Eligibility criteria (EC) are essential for clinical trial design, yet drafting them remains a time-intensive and cognitively demanding task for clinicians. Existing automated approaches often fall at two extremes—either requiring highly structured inputs, such as predefined entities to generate specific criteria, or relying on end-to-end systems that produce full eligibility criteria from minimal input such as trial descriptions—limiting their practical utility. In this work, we propose a guided generation framework that introduces interpretable semantic axes, such as \textit{Demographics}, \textit{Laboratory Parameters}, and \textit{Behavioral Factors}, to steer EC generation. These axes, derived using large language models, offer a middle ground between specificity and usability, enabling clinicians to guide generation without specifying exact entities. In addition, we present a reusable rubric-based evaluation framework that assesses generated criteria along clinically meaningful dimensions. Our results show that our guided generation approach consistently outperforms unguided generation in both automatic, rubric-based and clinician evaluations, offering a practical and interpretable solution for AI-assisted trial design.
\end{abstract}

\paragraph*{Data Availability} We used data that are publicly available at \url{https://clinicaltrials.gov}. 

\paragraph*{Institutional Review Board (IRB)}
This study uses publicly available, de-identified data from \url{https://clinicaltrials.gov} and therefore does not involve human subjects or require IRB approval.

\section{Introduction}
\label{sec:intro}
Eligibility criteria are a cornerstone of clinical trial design, defining the population for which an intervention is intended and ensuring both patient safety and scientific rigor. However, drafting comprehensive and contextually appropriate eligibility criteria remains a time-consuming and cognitively demanding task for clinicians and trial designers.
While clinicians are generally familiar with a core set of standard eligibility criteria \citep{denicoff2022implementing, ctep_guidance}—such as age, performance status, and common comorbidities—they often struggle to recall or articulate less frequent or more nuanced criteria. These less common criteria, though critical for refining the study population, are harder to recall and often require referencing prior trials or domain-specific guidelines.

To address this challenge, we propose \method (\textbf{P}rotocol \textbf{O}ptimization via \textbf{E}ligibility \textbf{T}uning), a framework for automatically generating eligibility criteria using large language models (LLMs), with a focus on a novel \textbf{guided generation} approach. In this setting, the model is not only given the trial metadata and existing criteria but is also directed to generate a new criterion within a specific semantic category or \textit{axis} (e.g., \textit{Demographics}, \textit{Laboratory and Clinical Parameters}, \textit{Behavioral and Lifestyle Factors}). These axes are themselves derived using LLMs and validated by experts, reducing manual effort while preserving clinical relevance. This dual benefit—minimizing clinician burden and improving generation quality—makes guided generation a practical and scalable solution for real-world clinical trial design.

Unlike approaches that require clinicians to specify exact entities for generation \citep{wang2022trial2vec}—an often impractical expectation given the complexity and variability of clinical concepts—guided generation allows clinicians to express intent at a higher level of abstraction. Clinicians may not recall precise entities but often have a general axis in mind along which they want to refine or expand criteria. Conversely, fully unguided generation from trial metadata may produce redundant or overly generic criteria that clinicians already know and do not need assistance with \citep{lekuthai-etal-2025-ec}. Guided generation strikes a balance by enabling targeted augmentation of eligibility criteria in a structured and interpretable manner.

To evaluate the quality of generated criteria, we go beyond standard automatic metrics such as BERTScore and introduce a rubric-based evaluation framework. This rubric assesses each generated criterion along multiple clinically meaningful dimensions. In addition to automatic and rubric-based evaluations, we conduct clinicians' assessments and find strong agreement between clinicians' judgments and those made by LLMs acting as evaluators. Notably, LLM-based evaluations closely mirror clinicians' assessments, suggesting that LLMs can serve as reliable proxies for expert review in scalable evaluation pipelines. Our results show that guided generation consistently outperforms unguided generation across both automatic and rubric-based evaluations, highlighting its potential to enhance the completeness, consistency, and efficiency of clinical trial design.

\begin{figure*}[!htb]
    \centering
    \includegraphics[width=1.0\linewidth]{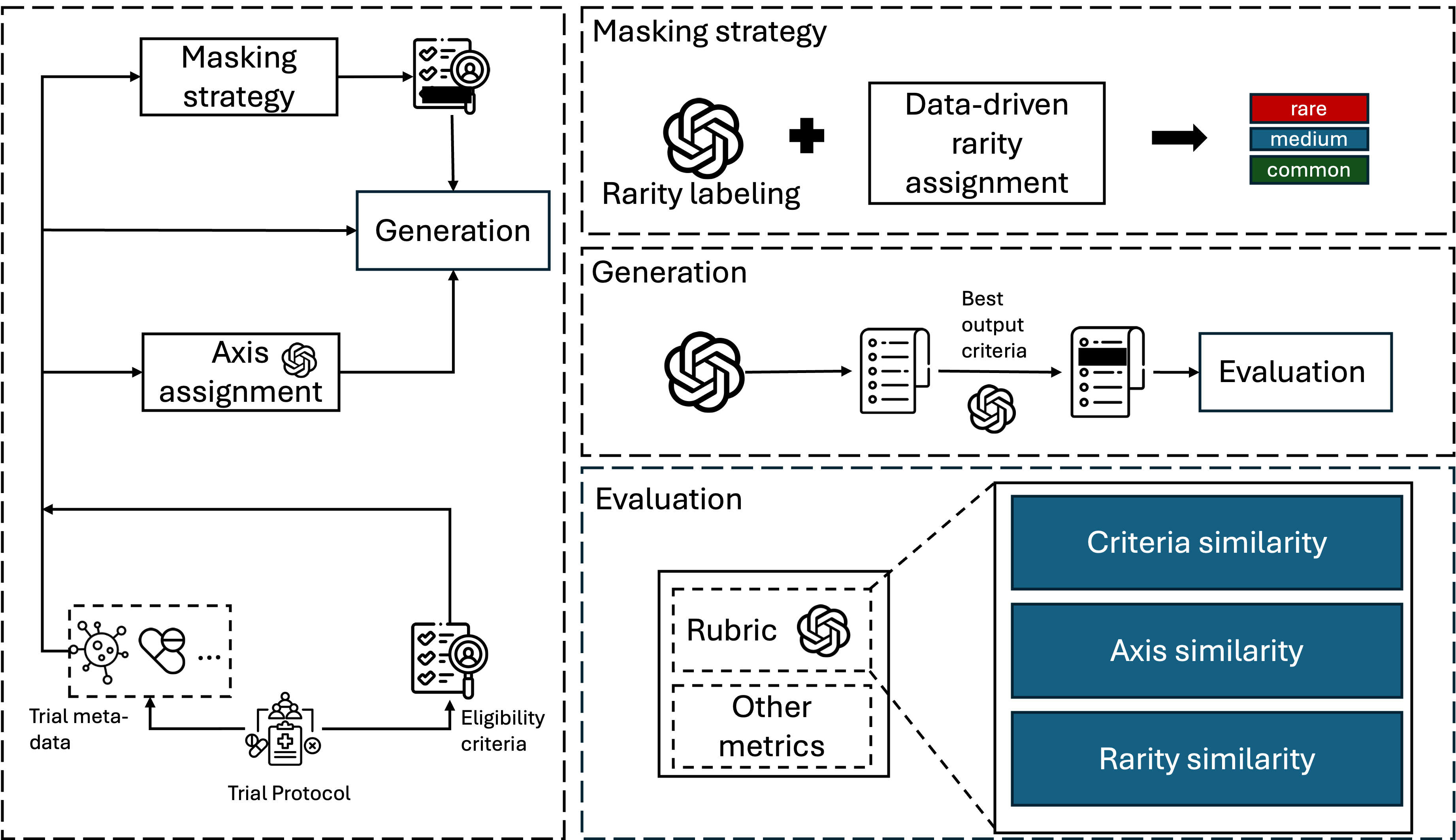}
    \caption{Framework of \method}
    \label{fig:enter-label}
\end{figure*}

\section{Related Work}
\paragraph{Eligibility Criteria Generation and Recommendation}
Recent work has explored the use of large language models (LLMs) and representation learning to support the design of clinical trial eligibility criteria (EC). \texttt{Trial2Vec} ~\citep{wang2022trial2vec} and \texttt{SECRET} ~\citep{das2025secret} generate trial-level representations using contrastive learning to recommend similar trials, laying the groundwork for EC retrieval. Building on this, \texttt{CReSE}~\citep{kim2024crese} applied contrastive learning and rephrasing strategies to recommend semantically relevant ECs for a given trial context.  \texttt{Autocriteria}~\citep{datta2024autocriteria} uses prompting on GPT-4 to extract granular ECs from trial documents, focusing on document-level extraction rather than generation. \texttt{AutoTrial}~\citep{wang2023autotrial} generates ECs using a hybrid prompting framework with LLMs, offering interpretability through explicit reasoning chains. However, it relies on fine-tuning and predefined entity-level categories, which may limit adaptability and usability in complex clinical settings. \texttt{EC-RAFT} is a recent method that generates complete EC directly from clinical trial titles and descriptions, without needing predefined EC categories or a recommendation system \citep{lekuthai-etal-2025-ec}. It uses retrieval-augmented fine-tuning to enhance its performance. In contrast, our approach uses LLM in a zero-shot setting to generate ECs guided by broader semantic axes—such as \textit{Demographics}, \textit{Laboratory Parameters}, and \textit{Behavioral Factors}. This enables interpretable and controllable generation without requiring fine-tuning or predefined entities, reducing clinician burden while maintaining flexibility and scalability for real-world clinical trial design.

\noindent\textbf{LLM Evaluation}
Large language models (LLMs) used in zero-shot or few-shot settings have demonstrated stronger alignment with human judgments compared to traditional lexical similarity metrics and earlier transformer-based embedding models across various natural language generation tasks ~\citep{fu2023gptscore, lin2024interpretable}. Moreover, studies have shown that LLMs can exhibit greater agreement with expert annotators than crowdworkers, further underscoring their reliability as evaluators \citep{gilardi2023chatgpt, chiang-lee-2023-large}. Rubric-based evaluation has emerged as a powerful alternative to traditional metrics for assessing LLM outputs in clinical and open-ended tasks ~\citep{hashemi2024llm, arora2025healthbench}. \texttt{HealthBench}~\citep{arora2025healthbench} is a recent benchmark that evaluates LLMs in healthcare using conversation-specific rubrics authored by physicians, covering over 48,000 unique criteria across 5,000 conversations. While HealthBench provides high-fidelity, clinically grounded evaluation, its per-example rubric design is resource-intensive and not easily scalable. In contrast, our rubric is reusable across examples and tailored to eligibility criteria generation, enabling structured, scalable evaluation along clinically meaningful dimensions.

\section{Method} \label{section: method}
\subsection{Problem Formulation}

Let a clinical trial be represented as a tuple:
\begin{equation}
T = (M, C)
\end{equation}
where:
\begin{itemize}
    \item $M$ denotes the trial metadata, including information such as condition, intervention, phase, and title.
    \item $C = \{c_1, c_2, \dots, c_n\}$ is the set of eligibility criteria, with each $c_i$ representing a single criterion.
\end{itemize}

We define a masking function $\mathcal{M}$ that removes one criterion $c_k$ from the set $C$, yielding a masked set:
\begin{equation}
C^{\setminus k} = \mathcal{M}(C, k) = \{c_1, \dots, c_{k-1}, \texttt{[MASK]}, c_{k+1}, \dots, c_n\}
\end{equation}

In addition to unguided generation, we introduce a guided generation setting where the model is prompted to generate $\hat{c}_k$ within a specific semantic category or \textit{axis}. These axes reflect common dimensions of eligibility criteria across trials and include: \textit{Demographics}, \textit{Diagnosis and Disease Characteristics}, \textit{Prior and Current Medical History}, \textit{Prior and Concomitant Treatments}, \textit{Laboratory and Clinical Parameters}, \textit{Performance and Functional Status}, \textit{Reproductive and Pregnancy Status}, \textit{Behavioral and Lifestyle Factors}, \textit{Infection and Immunization Status}, \textit{Consent, Legal, and Regulatory Compliance}, \textit{Administrative and Logistical Data}, \textit{Financial and Insurance Information}, and \textit{Other}.

The task is to utilize a language model $\mathcal{L}$ that, given the masked criteria set $C^{\setminus k}$, metadata $M$ and axis $A$, generates a prediction $\hat{c}_k$ for the missing criterion.

\begin{equation}
\hat{c}_k = \mathcal{L}(M, C^{\setminus k}, A)
\end{equation}
where $A$ is the target axis (e.g., \textit{Demographics}, \textit{Laboratory and Clinical Parameters}) that constrains the generation to a specific semantic category. Guided generation allows for more targeted assistance, enabling clinicians to request suggestions within a specific axis where they may need support. 

\subsection{Masking Strategy}

To simulate a realistic clinical trial design scenario—where clinicians are familiar with common eligibility criteria but may struggle to recall or articulate less frequent ones—we developed a masking strategy that can selectively target criteria from different rarity categories for generation (Figure \ref{fig:enter-label}). 

\subsubsection*{Step 1: Initial Labeling with LLM}

We employed an LLM to assign a preliminary label to each criterion $c_i \in C$ as one of the following given the disease name: common, medium (synonymous to medium-common), rare, invalid.

\subsubsection*{Step 2: Data-Driven Similarity Analysis}

To refine this classification, we adopted a data-driven approach using disease-specific corpora. For each disease $d$, we collected all associated trials and embedded each criterion $c_i$ using BioBERT to obtain a dense vector representation:
\begin{equation}
\mathbf{v}_i = \text{BioBERT}(c_i)
\end{equation}

We then computed the cosine similarity between each pair of criteria within the same disease:
\begin{equation}
\text{sim}(c_i, c_j) = \frac{\mathbf{v}_i \cdot \mathbf{v}_j}{\|\mathbf{v}_i\| \|\mathbf{v}_j\|}
\end{equation}

A similarity threshold $\theta_d$ was defined for each disease $d$ such that the most common criterion in that disease has at most as many similar instances as the number of trials $N_d$ for that disease:
\begin{equation}
\max_{c_i} \left| \left\{ c_j \mid \text{sim}(c_i, c_j) > \theta_d \right\} \right| \leq N_d
\end{equation}

This ensures that no criterion is counted more than once per trial, aligning with the assumption that each criterion appears only once per trial. For each criterion, we counted the number of similar criteria (above the threshold $\theta_d$) within the disease-specific dataset. Based on this count, we divided all criteria into deciles: lower (1--3, rare), middle (4--7, medium), and upper (8--10, common).

\subsubsection*{Step 3: Consensus-Based Filtering}

To ensure robustness in our zero-shot generation setup, we masked only those criteria that were consistently labeled as rare or medium or common by both the LLM and the data-driven approach. This consensus-based selection helps reduce noise and ambiguity in the benchmark and reflects realistic scenarios where clinicians are confident about common criteria but may need assistance with less frequent ones. 

\begin{figure}
    \centering
    \includegraphics[width=1.0\linewidth]{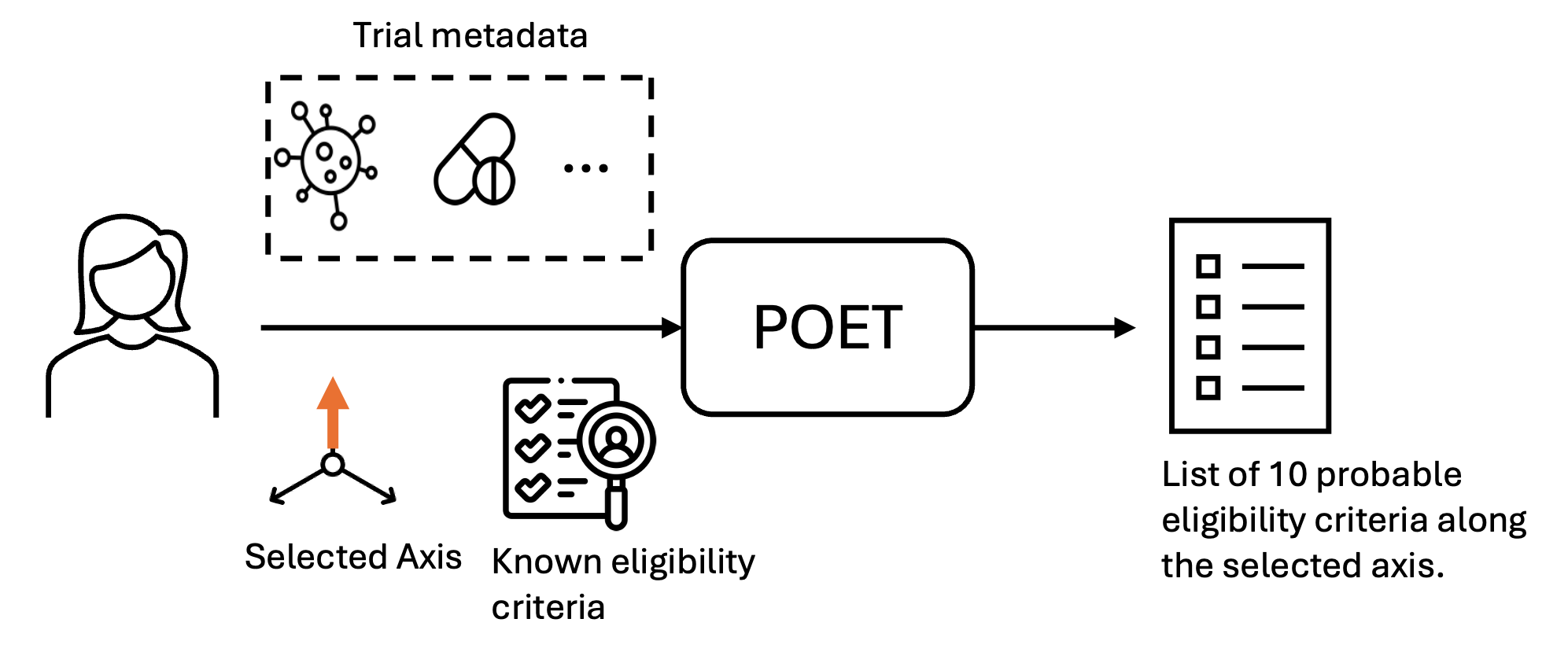}
    \caption{Usage of \method}
    \label{fig:inference}
    \vspace{-1em}
\end{figure}

Notably, our analysis (see Appendix, Figure \ref{fig:rarity}) shows that both labeling approaches exhibit similar trends. This suggests that researchers with limited resources can choose either method: those without access to large annotated datasets may rely on LLM-based estimation, while those without advanced models may use corpus statistics. However, for inference, clinicians do not need rarity labels to use \method (Figure \ref{fig:inference}). The masking strategy is only necessary for the development of our framework without clinician annotation for each criterion.

\subsection{Criteria Generation using LLM}

We use LLMs to generate missing eligibility criteria in a zero-shot setting. Given a clinical trial $T = (M, C)$ and axis $A$, where $M$ includes metadata such as disease, intervention, and title, and $C^{\setminus k}$ is the set of criteria with one masked element, the model is prompted to generate a plausible replacement $\hat{c}_k$.

\subsection{Rubric Creation}
To evaluate the quality of generated eligibility criteria, we define a rubric with three dimensions: \textit{Criteria similarity}, \textit{Axis similarity}, and \textit{Rarity similarity}. Each dimension is scored independently as follows:

\subsubsection*{Criteria Similarity (0--3)}
This score captures the semantic similarity between the generated criterion $\hat{c}_k$ and the ground truth criterion $c_k$. It is evaluated on a 4-point scale:

\begin{itemize}
    \item \textbf{0 (No similarity):} Completely different concepts or unrelated clinical meaning. \textit{Example:} Target: ``Patients with diabetes'' \quad Generated: ``Patients with asthma''
    
    \item \textbf{1 (Low similarity):} Some thematic overlap, but core clinical meaning differs significantly. \textit{Example:} Target: ``Stage 3 kidney disease'' \quad Generated: ``Any kidney condition''
    
    \item \textbf{2 (Moderate similarity):} Same general intent, with minor differences in specificity, scope, or terminology. \textit{Example:} Target: ``Patients with type 2 diabetes'' \quad Generated: ``Adults with diabetes mellitus type 2''
    
    \item \textbf{3 (High similarity / Equivalent):} Nearly identical meaning, including clinically equivalent expressions. \textit{Example:} Target: ``Karnofsky $\geq$ 70'' \quad Generated: ``ECOG 0 or 1''
\end{itemize}

\subsubsection*{Axis Similarity (0 or 1)}
This binary score assesses whether the generated criterion belongs to the same semantic axis as the target criterion. A score of 1 indicates correct axis alignment; 0 indicates a mismatch.

\subsubsection*{Rarity Similarity (0 or 1)}
This score evaluates whether the generated criterion matches the rarity category (e.g., rare, medium, common) of the target criterion. A score of 1 indicates correct rarity alignment; 0 indicates a mismatch.

\begin{table}[!htbp]
\centering
\caption{Dataset Statistics}
\label{tab:dataset_stats}
\begin{tabular}{ll}
\toprule
\textbf{Statistic} & \textbf{Value} \\
\midrule
Number of Criteria & 1235 \\
Number of Clinical Trials & 344 \\
\midrule
\multicolumn{2}{l}{\textbf{Diseases}} \\
Leukemia & 232 \\
Multiple Myeloma & 167 \\
Prostatic Neoplasms & 149 \\
Melanoma & 147 \\
Breast Neoplasms & 122 \\
Lung Neoplasms & 97 \\
HIV Infections & 96 \\
Diabetes Mellitus, Type 2 & 87 \\
Schizophrenia & 75 \\
Hypertension & 63 \\
\midrule
\multicolumn{2}{l}{\textbf{Rarity}} \\
Common & 736 \\
Medium & 396 \\
Rare & 103 \\
\bottomrule
\end{tabular}
\end{table}

\section{Experiments and Results}
\subsection{Experimental Setup}

\begin{table*}[!htb]
\centering
\caption{Comparative Evaluation of GPT-4.1 Performance in Guided vs. Unguided Settings Across Rarity Levels Using LLM-as-a-Judge and BERTScore Metrics. \textbf{Bold} numbers indicate better performance within each category (rare, medium, common, or overall). Values are reported as mean $\pm$ standard deviation.}
\label{tab:combined_results}
\begin{tabular}{llccc|c}
\toprule
\textbf{Rarity} & \textbf{Setting} & \multicolumn{3}{c|}{\textbf{LLM-as-a-Judge}} & \textbf{BERTScore} \\
\cmidrule(lr){3-5}
& & \textbf{Criteria} & \textbf{Axis} & \textbf{Rarity} & \\
\midrule
\multirow{2}{*}{Rare} 
  & Unguided & 0.15 $\pm$ 0.55 & 0.44 $\pm$ 0.50 & 0.14 $\pm$ 0.34 & 0.38 $\pm$ 0.08 \\
  & Guided   & \textbf{0.50 $\pm$ 0.71} & \textbf{0.99 $\pm$ 0.10} & \textbf{0.56 $\pm$ 0.50} & \textbf{0.45 $\pm$ 0.10} \\
\midrule
\multirow{2}{*}{Medium} 
  & Unguided & 0.28 $\pm$ 0.61 & 0.37 $\pm$ 0.48 & 0.76 $\pm$ 0.43 & 0.40 $\pm$ 0.10 \\
  & Guided   & \textbf{0.93 $\pm$ 0.68} & \textbf{1.00 $\pm$ 0.05} & \textbf{1.00 $\pm$ 0.00} & \textbf{0.51 $\pm$ 0.10} \\
\midrule
\multirow{2}{*}{Common} 
  & Unguided & 0.73 $\pm$ 1.06 & 0.50 $\pm$ 0.50 & 0.99 $\pm$ 0.08 & 0.48 $\pm$ 0.16 \\
  & Guided   & \textbf{1.77 $\pm$ 0.89} & \textbf{0.99 $\pm$ 0.07} & \textbf{1.00 $\pm$ 0.05} & \textbf{0.62 $\pm$ 0.14} \\
\midrule
\multirow{2}{*}{All} 
  & Unguided & 0.54 $\pm$ 0.93 & 0.45 $\pm$ 0.50 & 0.85 $\pm$ 0.36 & 0.45 $\pm$ 0.14 \\
  & Guided   & \textbf{1.39 $\pm$ 0.94} & \textbf{1.00 $\pm$ 0.07} & \textbf{0.96 $\pm$ 0.19} & \textbf{0.57 $\pm$ 0.14} \\
\bottomrule
\end{tabular}
\end{table*}

We used a publicly available clinical trial dataset from \url{https://aact.ctti-clinicaltrials.org}, filtering for U.S.-based interventional drug trials. Trials were selected based on 10 common diseases, standardized using MeSH terms (see Table~\ref{tab:dataset_stats}). From 400 trials containing 3658 criteria, we extracted 1235 inclusion criteria from 344 trials using a consensus-based filtering approach. Each criterion was assigned a rarity label via LLM and data-driven similarity analysis, and categorized along semantic axes using LLMs (see Section~\ref{section: method} for details).

We evaluated generation quality using both BERTScore~\citep{zhang2019bertscore} and an LLM-as-a-Judge framework based on our rubric. Experiments were conducted using various GPT models (GPT-4.1, GPT-4o, GPT-4o-mini, o3, o1), with a temperature set to 0.0 for deterministic results. All generations were performed via the Azure OpenAI API for enhanced security. The experiments were run on an Azure Standard\_E4s\_v3 VM with 4 vCPUs, 32 GB RAM, and 64 GB disk space.

\subsection{Results and Discussion}
\subsubsection{Experiments on Guided vs Unguided Generation}
We asked GPT 4.1 to suggest 10 probable criteria for a target/masked criteria given other criteria and clinical trial metadata like title, disease name, intervention, etc. We discuss the results in the next section.

To assess the effectiveness of our guided approach, we compare it against the unguided setting across rare, medium, common, and overall criteria, as shown in Table~\ref{tab:combined_results}. All results are based on the \textit{first suggested output} generated by GPT-4.1 and evaluated against the target criteria. Guided generations consistently outperform unguided ones across all rarity levels.  Improvements range from approximately 18--300\% depending on the metric and rarity level, with particularly large gains for rare and medium criteria. For common criteria, the gains are more modest but remain consistently positive. Statistical analysis confirms that the guided framework provides highly significant performance gains ($p < 0.05$) over the unguided version across nearly all criteria types and metrics (Table \ref{tab:statistical-tests}). Addition results comparing EC-RAFT with \method is available in the Appendix (Table \ref{tab:ec-raft-comparison}).

\begin{figure}[!htb]
    \centering
    \includegraphics[width= 1.0\linewidth]{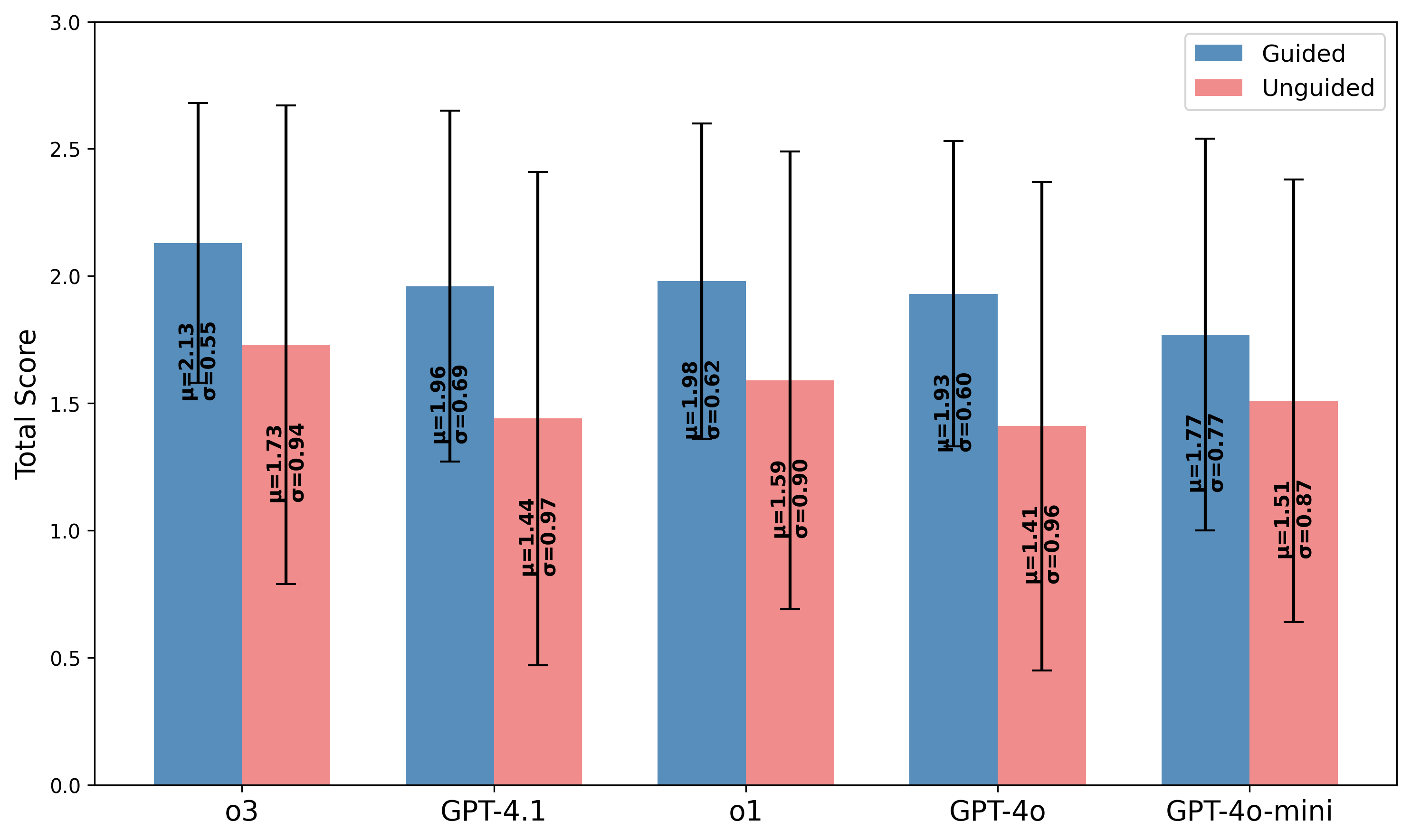} 
    \caption{Comparison among different GPT models on rare data. \textit{Total score} is computed as the sum of three components: \textit{Criteria Similarity} (normalized between 0 and 1), \textit{Axis Similarity}, and \textit{Rarity Similarity}. Values are reported as mean $\pm$ standard deviation.}
    \label{fig:baselines}
    \vspace{-1em}
\end{figure}

\begin{figure*}[!htb]
    \centering    \includegraphics[width=\textwidth]{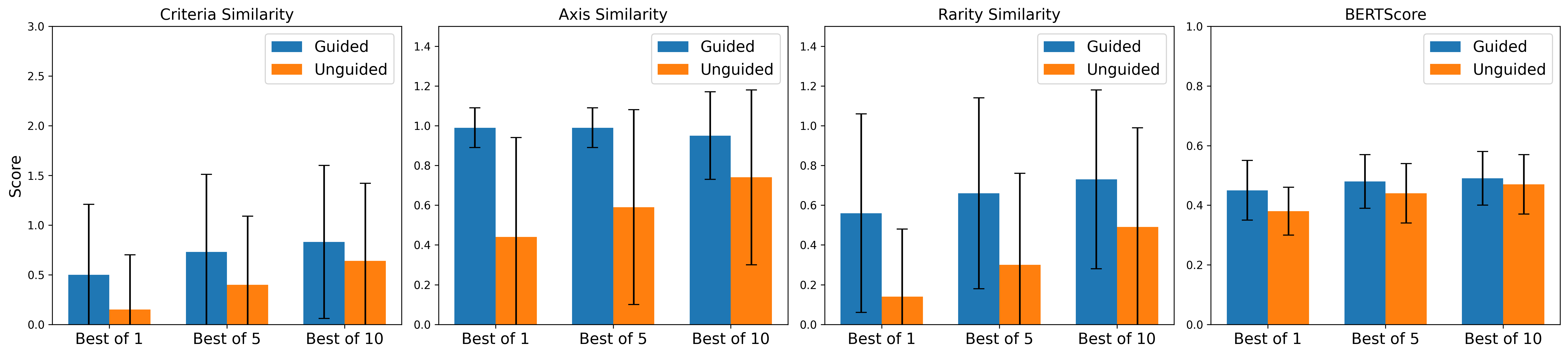}
    \caption{Comparison of guided and unguided approaches across four metrics—\textit{Criteria Similarity} (range: 0–3), \textit{Axis Similarity} (binary: 0 or 1), \textit{Rarity Similarity} (binary: 0 or 1), and \textit{BERTScore} (continuous, between 0 and 1)—for Best of 1, 5, and 10 generations compared to corresponding target criteria in rare category. Error bars represent standard deviation.}
    \label{fig:rare}
    \vspace{-1em}
\end{figure*}

We also show on different baselines that guided is better than unguided generations on the rare test set. Figure~\ref{fig:baselines} presents a comparative analysis of five GPT models—o3, GPT-4.1, o1, GPT-4o, and GPT-4o-mini—under guided and unguided generation settings for rare scenario eligibility criteria. Across all models, the guided setting consistently outperforms the unguided setting, indicating that semantic guidance significantly enhances the relevance and completeness of generated criteria. o3 achieves the highest guided score ($\mu=2.13$, $\sigma=0.55$), while GPT-4o-mini shows the lowest guided score ($\mu=1.77$, $\sigma=0.77$), suggesting that model size and architecture may influence performance in rare data scenarios. These results are on the best out of 10 generated outputs compared with the corresponding ground truth criterion.

\subsubsection{Experiments of Number of Generation}
In this section, we present results for GPT-4.1 using the best generated criterion per target criterion, selected from the top 1, 5, and 10 outputs. As a reminder, the model was prompted to generate 10 candidate eligibility criteria for each masked (target) criterion, using the remaining criteria and clinical trial metadata (e.g., title, disease name, intervention) as context. From these 10 outputs, we identified the most similar criterion among the top 1, 5, and 10 candidates using GPT-4.1.

\begin{figure}[!htb]
    \centering
    \includegraphics[width=0.95\linewidth]{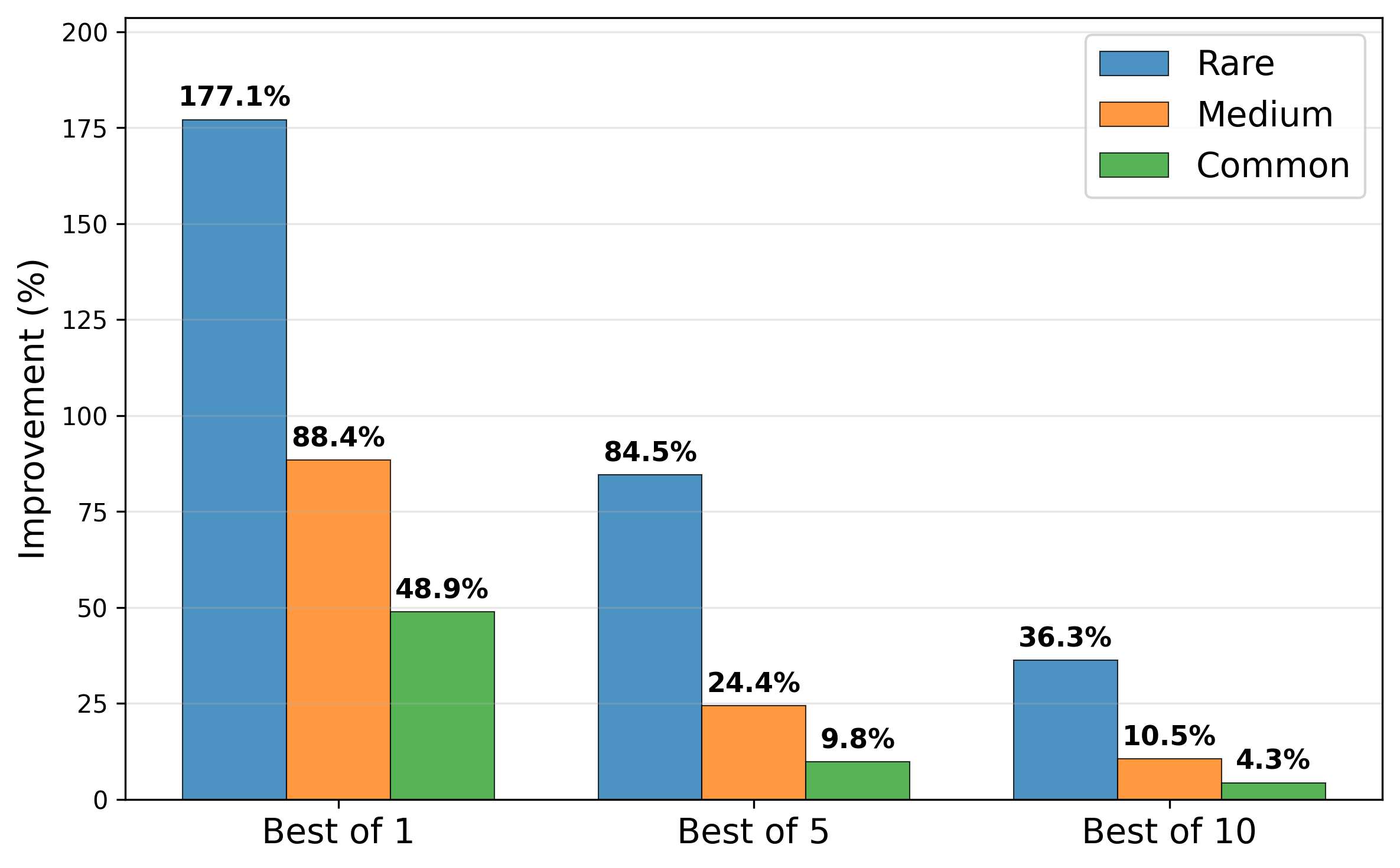}
    \caption{Improvement in mean \textit{Total Score} for guided over unguided generation across rare, medium, and common categories. }
    \label{fig:improvement}
    \vspace{-1em}
\end{figure}

Across all rarity levels, increasing the number of generated outputs (from 1 to 10) improves performance for both guided and unguided settings. However, guided generation consistently shows stronger and more stable improvements. The trends from Figure \ref{fig:rare} suggests that generating more candidate outputs helps improve quality, and guided generation benefits more consistently from this increase for rare criteria.
Figure~\ref{fig:improvement} illustrates the percentage improvement in mean \textit{Total Score} for guided generation over unguided generation across rare, medium, and common categories, evaluated under three sampling configurations: Best of 1, 5, and 10. The results show that guided generation yields the most substantial gains in the rare category, with a 177.1\% improvement in the Best of 1 setting. As the number of samples increases, the improvement diminishes across all categories, suggesting that, while sampling helps refine outputs, the relative advantage of guidance is most pronounced when fewer generations are available. The common category shows the least improvement, indicating that guidance is particularly beneficial for generating criteria in low-frequency or complex scenarios. Here, \textit{Total Score} is calculated as the sum of normalized \textit{Criteria Similarity} (0–1), \textit{Axis Similarity} (0 or 1), and \textit{Rarity Similarity} (0 or 1).

\begin{table*}[!htb]
\centering
\resizebox{\textwidth}{!}{%
\begin{tabular}{|l|p{2.5cm}|p{3cm}|p{3.5cm}|p{3.5cm}|p{4.5cm}|p{4.5cm}|}
\hline
\textbf{NCTID} & \textbf{Guide} & \textbf{Target Criterion} & \textbf{Guided (Best of 1)} & \textbf{Guided (Best of 10)} & \textbf{Unguided (Best of 1)} & \textbf{Unguided (Best of 10)} \\
\hline
NCT00082394 & Laboratory and Clinical Parameters & Plasma HIV-1 RNA between 500 and 20,000 copies/mL & Plasma HIV-1 RNA (viral load) greater than 1,000 copies/mL at screening & Plasma HIV-1 RNA (viral load) greater than 1,000 copies/mL at screening & Age 18 years or older at the time of screening & Plasma HIV-1 RNA (viral load) greater than 1,000 copies/mL at screening \\
\hline
NCT00017719 & Laboratory and Clinical Parameters & CD4 count less than 250 cells/mm\textsuperscript{3} within 30 days of study entry & CD4+ T-cell count greater than 100 cells/mm\textsuperscript{3} within 30 days prior to study entry & CD4+ T-cell count greater than 100 cells/mm\textsuperscript{3} within 30 days prior to study entry & Age 18 years or older at the time of enrollment & Hemoglobin level of at least 8.0 g/dL within 30 days prior to study entry \\
\hline
\end{tabular}%
}
\caption{Comparison of Target vs. Generated Eligibility Criteria in Guided and Unguided Settings for Two Clinical Trials}
\label{tab:criteria_comparison}
\end{table*}

\subsection{Clinician Evaluation}
We conducted a human evaluation by clinicians using 100 target eligibility criteria, selected to reflect the original dataset's distribution of rare, medium, and common criteria. For each target, clinicians were presented with both guided and unguided suggestions in a blinded setup. Two expert annotators rated the similarity of each suggestion on a scale from 0 to 3, following the \textit{Criteria Similarity} rubric.

\begin{figure}
    \centering
    \includegraphics[width=0.98\linewidth]{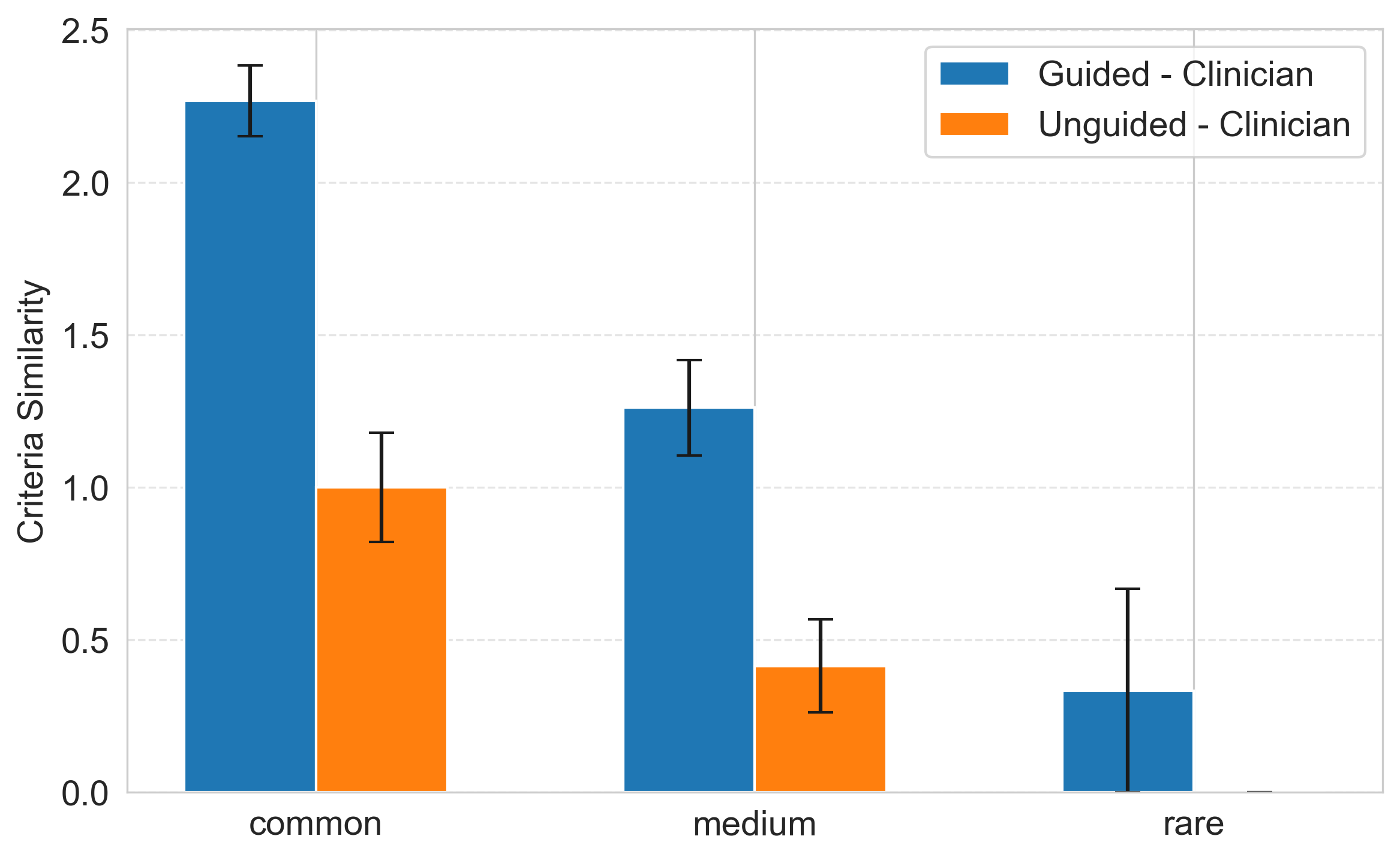}
    \caption{Clinicians' evaluation for \textit{Criteria Similarity} (0-3) according to the rubric. Error bars represent standard deviation.}
    \label{fig:human_guided_vs_unguided}
    \vspace{-1em}
\end{figure}

\begin{figure}[!htb]
    \centering
    \includegraphics[width=0.9\linewidth]{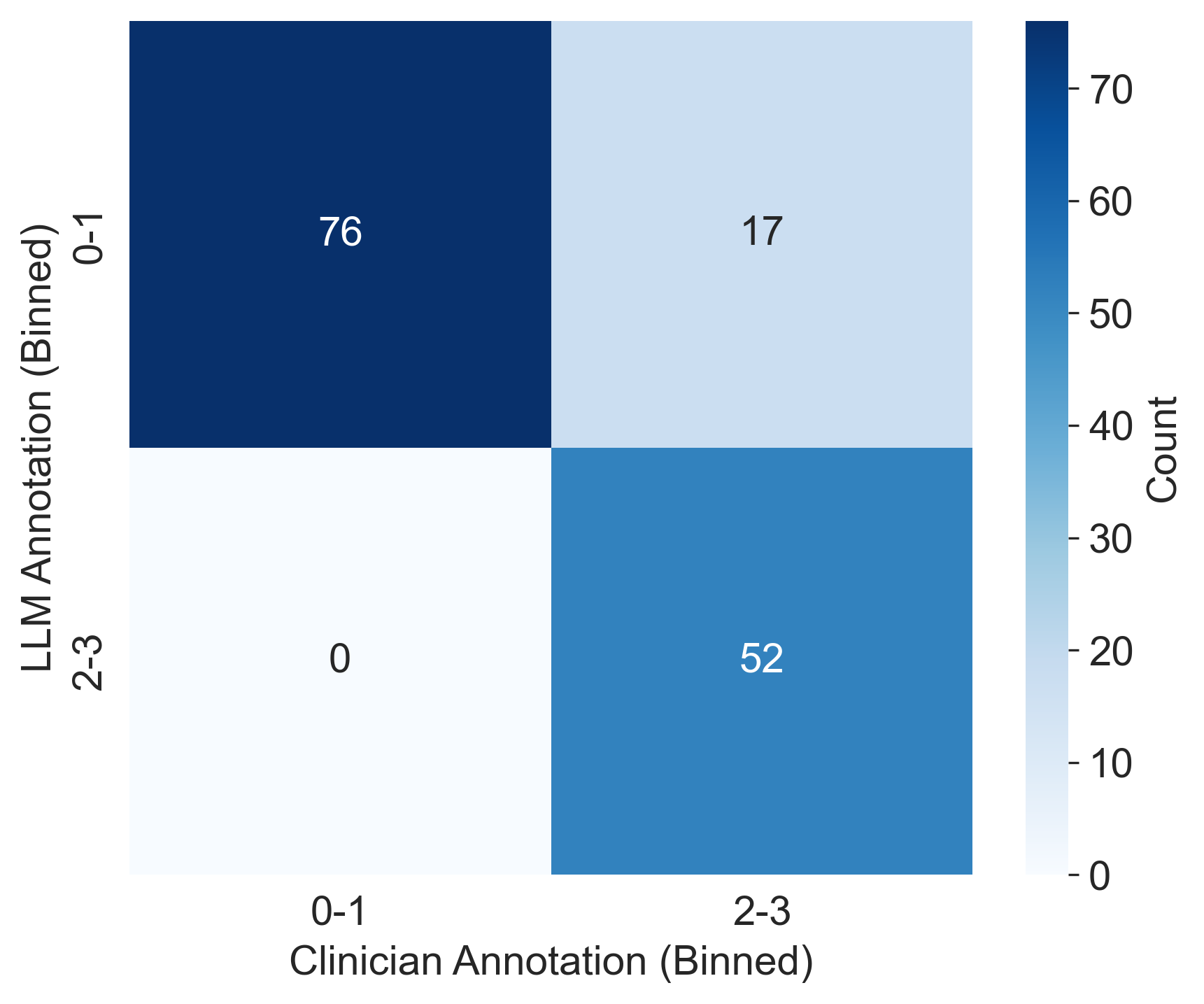}
    \caption{Confusion matrix illustrating agreement between LLM and Clinician annotations after binning. Accuracy score 88.3\%.}
    \label{fig:human}
    \vspace{-2em}
\end{figure}

Out of 200 total ratings, we retained 145 samples where annotators agreed on their scores. These scores are visualized in Figure~\ref{fig:human_guided_vs_unguided}, which shows trends consistent with LLM-as-a-Judge evaluations (see Figures~\ref{fig:rare}, \ref{fig:medium}, and \ref{fig:common}). Figure~\ref{fig:human} compares clinician and LLM judgments, revealing that the LLM scores and clinician scores have significant overlaps. The agreement rate between clinician and LLM judgments was $77.24\%$ ($112/145$) (Figure \ref{fig:tolerance}). If we bin the scores into 0-1 and 2-3 groups, we get an accuracy of 88.3\% indicating that LLMs can serve as reliable proxies for expert evaluation under relaxed condition (Figure \ref{fig:human}). %. Allowing a tolerance of $\pm 1$ increases the agreement rate to $99.3\%$, 

\subsection{Case Studies}

For the case studies, we selected two clinical trials (NCT00017719 and NCT00082394) from \url{https://clinicaltrials.gov}, each of which included at least one eligibility criterion that was modified during the course of the clinical trial. 
% We assume that clinicians were already aware of the eligibility criteria listed in the first version of the protocol. 
To illustrate the use of \method, we assumed that modified criteria were suggested by experts or derived from external knowledge sources, including existing clinical trials. 
% We hope to help the clinicians at this stage to get suggestions for additional criteria using \method (Figure \ref{fig:inference}). 
% \method helps clinicians at this stage to derive suggestions for additional criteria (Figure \ref{fig:inference}).
Table~\ref{tab:criteria_comparison} shows that guided generation yields the best matches to the target criteria in the first output among the ten generated outputs. For NCT00082394, the best match from the ten unguided generations is identical to the first output from the guided generation. In contrast, for NCT00017719, the best match from the ten unguided generations does not closely resemble the target criterion, whereas the first guided output provides the best match.

\section{Conclusion}
Eligibility criteria are essential to clinical trial design, yet their creation remains a labor-intensive and cognitively demanding process. In this work, we introduced \texttt{POET}, a framework for generating eligibility criteria using large language models, with a novel guided generation approach that leverages semantic axes to improve generation. By automatically deriving these axes and directing generation accordingly, \method reduces clinician burden while enhancing the specificity and completeness of trial criteria.
Our rubric-based evaluation framework, alongside standard metrics, demonstrates that guided generation consistently outperforms unguided approach across multiple dimensions. These results highlight the potential of guided LLM-based generation to support scalable, high-quality clinical trial design. 

%\section{Author Contributions}
%This section will be completed after %anonymous peer review.

\bibliography{chil-sample}

\newpage
\newpage
\appendix
\section{First Appendix}\label{apd:first}
\subsection{Prompting LLM for Generation}
We used the prompts below for all the models to generate 10 probable/reasonable criteria given all other information. Guided generation has the information of axis for each target criterion whereas unguided generation does not have that information. All other pieces of information are exactly the same for two approaches.

\tcbset{colback=gray!10, colframe=gray!80, boxrule=0.5pt, arc=2pt, left=2pt, right=2pt, top=2pt, bottom=2pt}

\begin{tcolorbox}[title=Prompt for Guided Generation]
The first few inclusion criterion for a clinical trial are (separated by $|$): \{input\_criteria\}. 
The condition/disease investigated in this trial is \{disease\}. 
The intervention investigated in this trial is \{intervention\_name\}. 
The phase of this trial is \{phase\}. 
The primary outcome measure/s of this trial is/are \{primary\_outcome\_measure\}. 
The title of this trial is \{brief\_title\}. 
Focus on generating criteria specifically related to the following axis: \{assigned\_axis\}. 
Based on this information, generate a reasonable set of 10 more inclusion criteria for this trial. Do not repeat those already in the input. 
Output each criterion as a separate line, and wrap all generated criteria in $<$final\_answer$>$ ... $<$/final\_answer$>$ tags. 
Do not output anything else (i.e., any pretext or post text) other than the generated criteria.
\end{tcolorbox}
\tcbset{colback=gray!10, colframe=gray!80, boxrule=0.5pt, arc=2pt, left=2pt, right=2pt, top=2pt, bottom=2pt}
\begin{tcolorbox}[title=Prompt for Unguided Generation]
The first few inclusion criterion for a clinical trial are (separated by $|$): \{input\_criteria\}. 
The condition/disease investigated in this trial is \{disease\}. 
The intervention investigated in this trial is \{intervention\_name\}. 
The phase of this trial is \{phase\}. 
The primary outcome measure/s of this trial is/are \{primary\_outcome\_measure\}. 
The title of this trial is \{brief\_title\}. 
Based on this information, generate a reasonable set of 10 more inclusion criteria for this trial. Do not repeat those already in the input. 
Output each criterion as a separate line, and wrap all generated criteria in $<$final\_answer$>$ ... $<$/final\_answer$>$ tags. 
Do not output anything else (i.e., any pretext or post text) other than the generated criteria.
\end{tcolorbox}

\subsection{Prompting for LLM-as-a-Judge}
\tcbset{colback=gray!10, colframe=gray!80, boxrule=0.5pt, arc=2pt, left=2pt, right=2pt, top=2pt, bottom=2pt}
\begin{tcolorbox}[title=Prompt for LLM-as-a-Judge Evaluation, breakable]
Given the following: Trial title: \{brief\_title\} 

Trial disease: \{disease\} 

Trial intervention: \{intervention\_name\} 

True criterion: \{target\_criterion\} 

Generated criterion: \{generated\_criterion\} 

Target axis: \{assigned\_axis\}  

Target rarity: \{llm\_label\} . Evaluate the following between the given True and Generated criterion:

\textbf{Criteria Similarity (scale 0–3):}  
0 – No similarity: Completely different concepts or unrelated clinical meaning.  

1 – Low similarity: Some thematic overlap, but core clinical meaning differs significantly.

2 – Moderate similarity: Same general intent, minor differences in specificity, scope, or terminology. 

3 – High similarity / Equivalent: identical/ nearly identical meaning, including clinically equivalent scales.  

\textbf{Axis Similarity (0 or 1):}  

1 – Generated criterion matches target axis.
0 – Axis does not match.  

\textbf{Rarity (0 or 1):}  

1 – Generated criterion matches target rarity category.  
0 – Does not match.  

Respond in the format:  

Similarity: $<$number$>$  

Axis similarity: $<$0 or 1$>$ 

Rarity: $<$0 or 1$>$  

Justification: $<$your explanation$>$
\end{tcolorbox}

\subsection{Additional Results}
\begin{figure}[!htb]
    \centering
    \includegraphics[width= 0.9\linewidth]{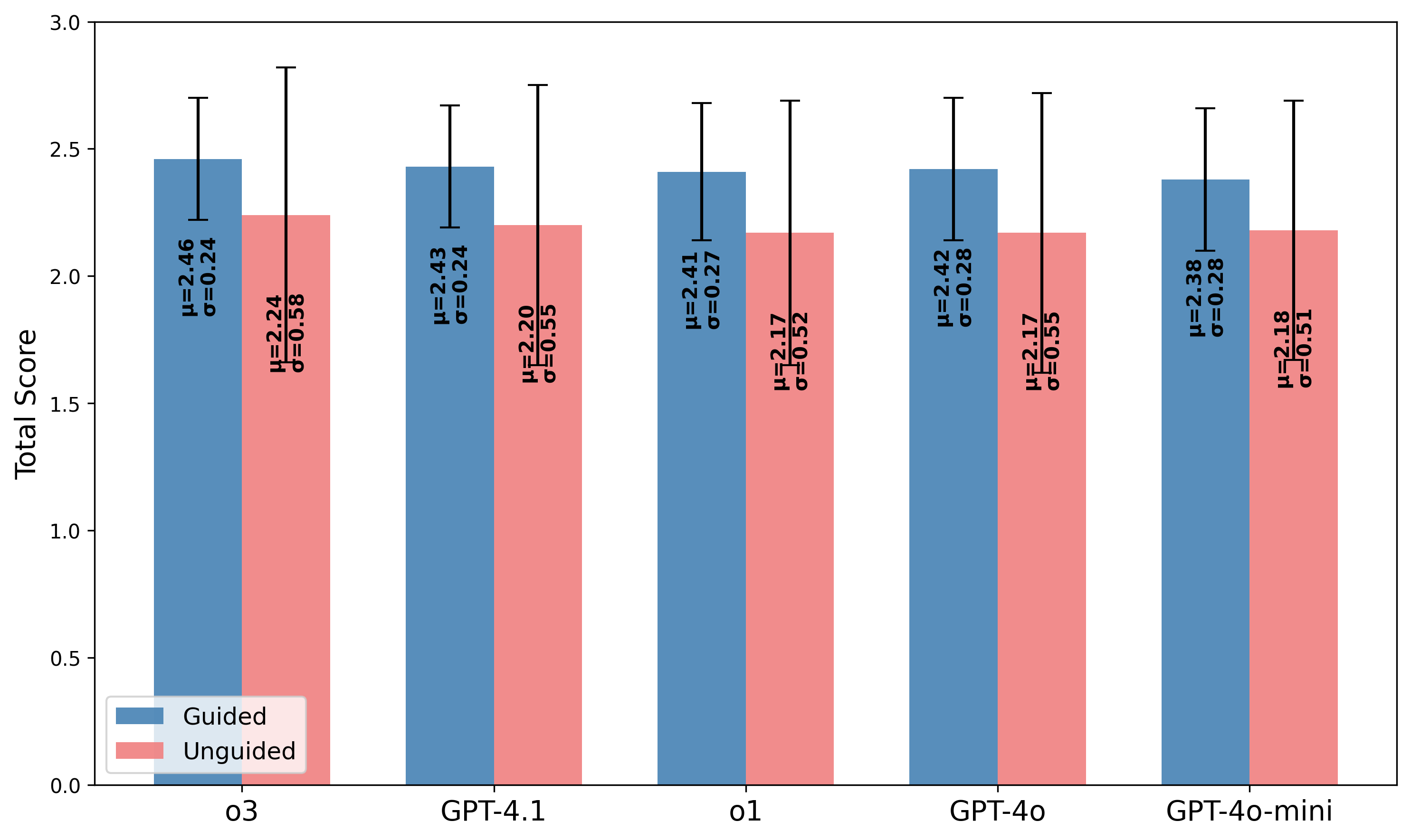} 
    \caption{Comparison among different GPT models on medium data. \textit{Total score} is computed as the sum of three components: \textit{Criteria Similarity} (normalized between 0 and 1), \textit{Axis Similarity}, and \textit{Rarity Similarity}.}
    \label{fig:baselines_medium}
\end{figure}

Figure \ref{fig:baselines_medium} shows comparison among different GPT models
on medium criteria. We only show results on medium (Figure \ref{fig:baselines_medium}) and rare (Figure \ref{fig:baselines}) criteria  as these are the situations where the clinicians need help during protocol designing. We observe that for rare criteria, the standard deviations of the \textit{Total Score} across different baselines in both guided and unguided scenarios are comparatively higher than those for medium criteria. Furthermore, when comparing guided and unguided scenarios, guided generations achieve higher scores and exhibit smaller standard deviations in both rare and medium cases. Notably, the improvements from guided over unguided generations are more pronounced in rare cases, establishing that guiding with axes can indeed assist clinicians during protocol optimization.

\begin{figure}[!htb]
    \centering
    \includegraphics[width=0.9\linewidth]{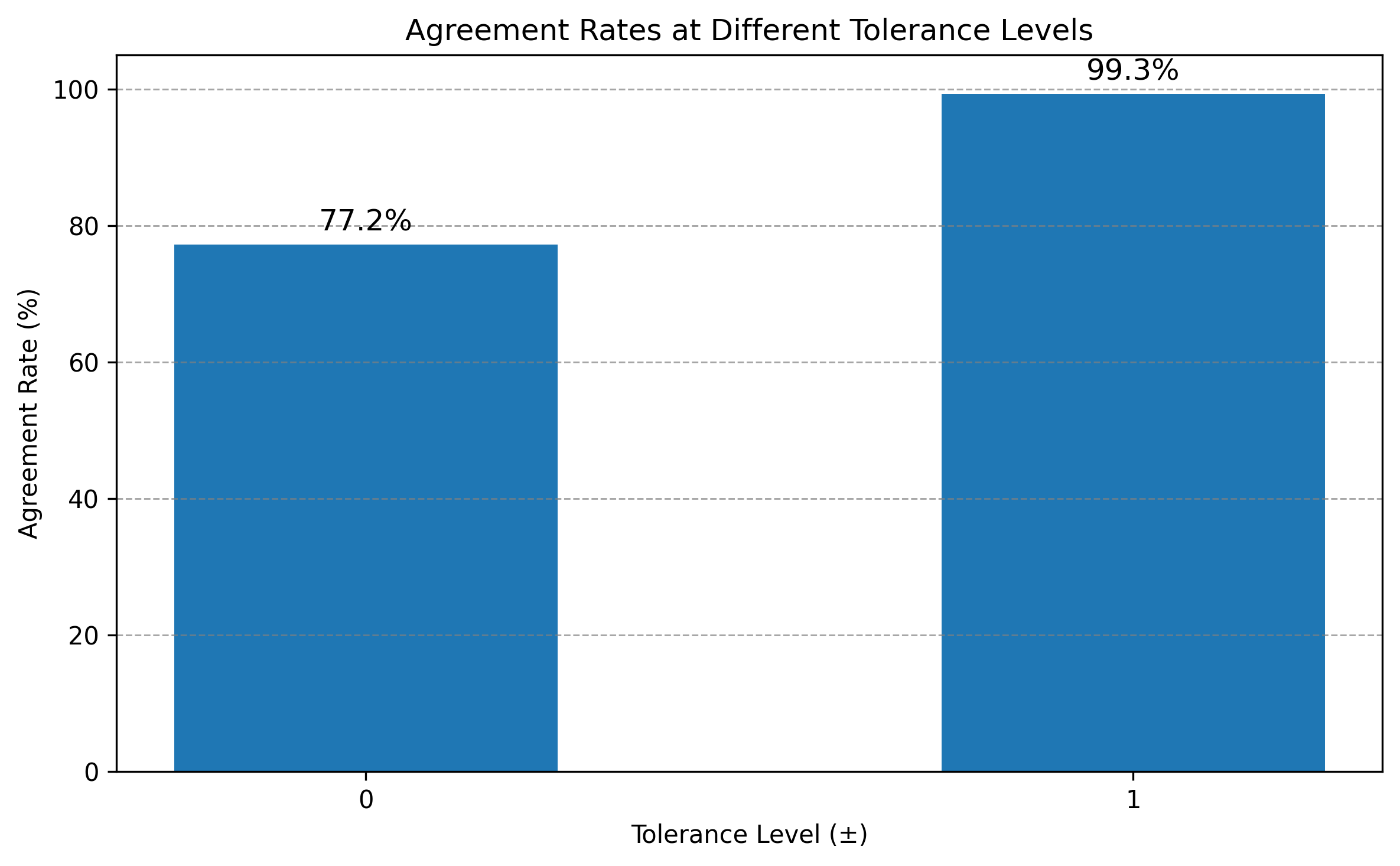}
    \caption{Agreement rates between LLM and Clinician scores at varying tolerance levels ($\pm 0$ and $\pm 1$): Higher tolerance shows increased alignment}
    \label{fig:tolerance}
\end{figure}

Figure \ref{fig:tolerance} presents the agreement rates between clinician and LLM scores across different tolerance levels. A manual review of the scores reveals that LLMs tend to grade more strictly than clinicians, even when applying the same rubric. Consequently, we observe that with a $\pm 1$ tolerance, GPT-4.1 Judge performs on par with clinicians, while at zero tolerance it is noticeably stricter.

\begin{figure*}[!htb]
    \centering
    \includegraphics[width=0.9\linewidth]{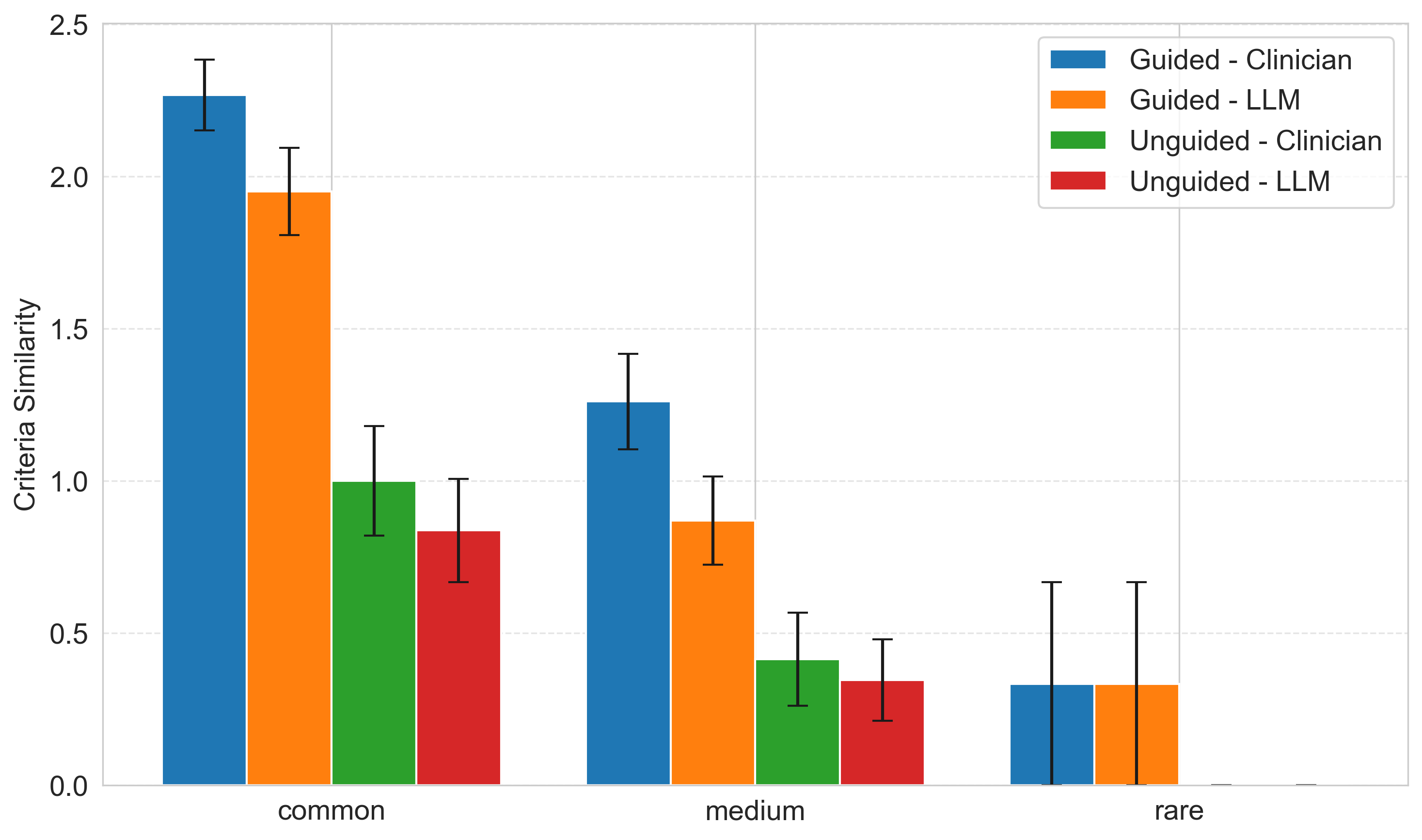}
    \caption{Clinician vs LLM judges across rare, medium and common categories.}
    \label{fig:human_vs_llm}
\end{figure*}

Figure \ref{fig:human_vs_llm} shows that guided generations based on high-level axes are judged more favorably by both LLMs and clinicians compared to unguided generations.

\begin{figure*}[!h]
    \centering
    \includegraphics[width=0.9\textwidth]{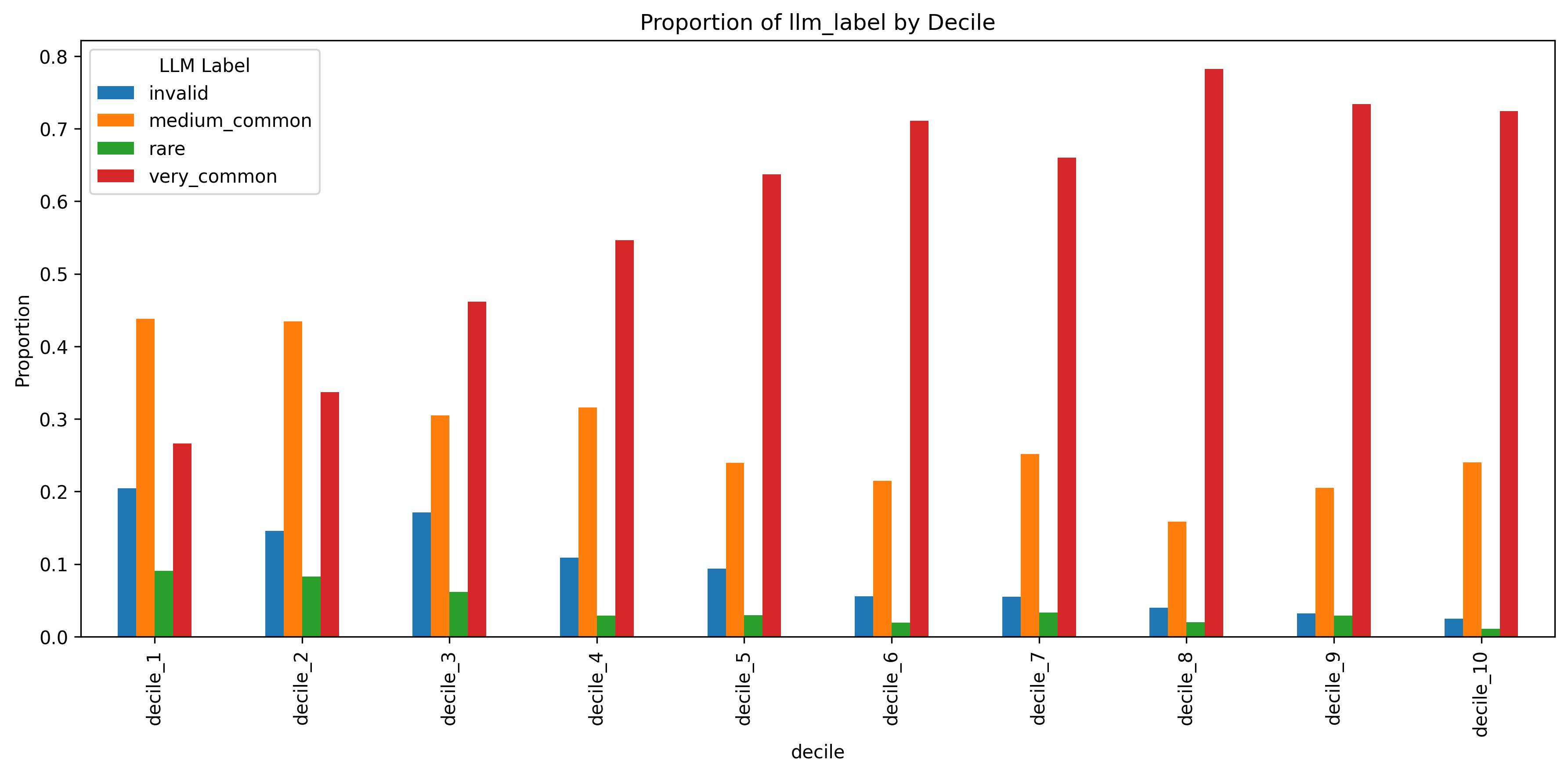}
    \caption{Rarity labels: LLM vs Data driven}
    \label{fig:rarity}
\end{figure*}

Figure \ref{fig:rarity} shows that there is similar trend between LLM labels and data-driven approach for rarity labeling with some exceptions. Therefore, we took a consensus-based approach to select the dataset in which both labeling approaches have the same labels for the data points. 

\begin{figure*}[!h]
    \centering
    \includegraphics[width=\textwidth]{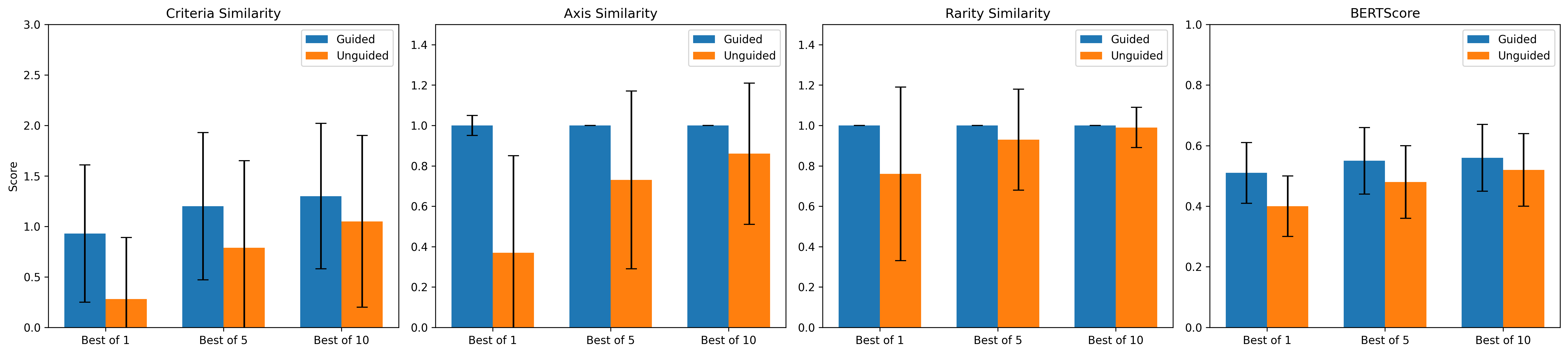}
    \caption{Evaluation metrics for medium scenario generation.  Error bars represent standard deviation.}
    \label{fig:medium}
\end{figure*}

\begin{figure*}[!h]
    \centering
    \includegraphics[width=\textwidth]{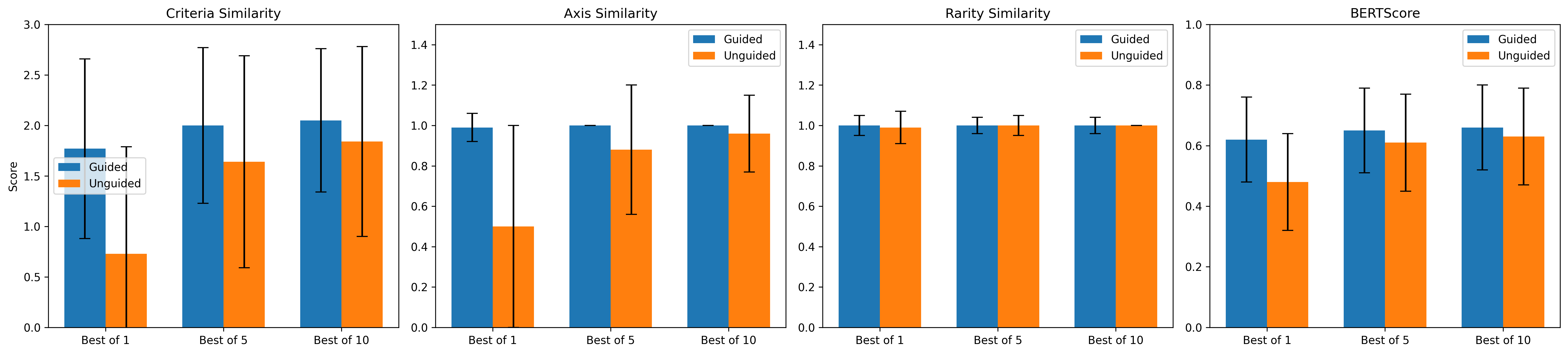}
    \caption{Evaluation metrics for common scenario generation. Error bars represent standard deviation.}
    \label{fig:common}
\end{figure*}

Figures \ref{fig:medium} and \ref{fig:common} exhibit trends consistent with Figure \ref{fig:rare}, further confirming that guided generation outperforms unguided generation. In general, we observe that guidance offers the greatest benefits in rare and medium cases, aligning with our expectations. However, in all cases, the incorporation of high-level axes into generation consistently improves both automatic metrics and rubric-based evaluations using LLM-as-a-Judge.

We ran EC-RAFT and evaluated its outputs using BERTScore. The results are shown in Table \ref{tab:ec-raft-comparison}. The other related work, AutoTrial, similarly does not have publicly available code or data, making it non-reproducible.

\begin{table}[ht]
\centering
\caption{Comparison of BERTScore between EC-RAFT and POET across different criteria types.}
\label{tab:ec-raft-comparison}
\begin{tabular}{lccc}
\hline
\textbf{Criteria Type} & \textbf{n} & \textbf{EC-RAFT} & \textbf{POET} \\ \hline
Rare                   & 5          & $0.13 \pm 0.07$  & $0.10 \pm 0.04$ \\
Medium                 & 27         & $0.12 \pm 0.12$  & $0.16 \pm 0.10$ \\
Common                 & 43         & $0.11 \pm 0.10$  & $0.21 \pm 0.10$ \\ \hline
All                    & 75         & $0.12 \pm 0.11$  & $0.18 \pm 0.10$ \\ \hline
\end{tabular}
\end{table}

Note that we only have 75 criteria in this test set to ensure that these trials are not part of EC-RAFT’s training data. There are only 5 rare samples in this set. We observe that EC-RAFT scores higher than POET on these rare samples, likely due to the retrieval-augmentation method in EC-RAFT. However, this difference on rare trials should be interpreted with caution, given the very small sample size. For medium, common, and all cases, POET outperforms EC-RAFT.

Statistical analysis confirms that the guided framework provides highly significant performance gains ($p < 0.05$) over the unguided version across nearly all criteria types and metrics \ref{tab:statistical-tests}. While improvements in semantic and axis similarity are robust for all groups, rarity categorization for common samples was the only metric that did not show a statistically significant change. Overall, the extreme significance levels ($p < 10^{-100}$) for "All Trials" demonstrate that the guidance method consistently and reliably improves model outputs.

\begin{table*}[ht]
\centering
\caption{Paired statistical tests for guided vs. unguided scores across criteria types.}
\label{tab:statistical-tests}
\begin{tabular}{lllc}
\hline
\textbf{Criteria Type} & \textbf{Metric (Test Type)} & \textbf{$p$-value} & \textbf{Sig.} \\ \hline
Rare & Criteria Similarity ($t$-test) & $6.171 \times 10^{-8}$ & Yes \\
Rare & Criteria Similarity (Wilcoxon) & $4.094 \times 10^{-7}$ & Yes \\
Rare & Axis Similarity (McNemar) & $2.082 \times 10^{-16}$ & Yes \\
Rare & Rarity Similarity (McNemar) & $1.307 \times 10^{-10}$ & Yes \\ \hline
Medium & Criteria Similarity ($t$-test) & $1.133 \times 10^{-51}$ & Yes \\
Medium & Criteria Similarity (Wilcoxon) & $1.754 \times 10^{-39}$ & Yes \\
Medium & Axis Similarity (McNemar) & $4.422 \times 10^{-75}$ & Yes \\
Medium & Rarity Similarity (McNemar) & $1.010 \times 10^{-28}$ & Yes \\ \hline
Common & Criteria Similarity ($t$-test) & $4.735 \times 10^{-97}$ & Yes \\
Common & Criteria Similarity (Wilcoxon) & $2.816 \times 10^{-69}$ & Yes \\
Common & Axis Similarity (McNemar) & $6.653 \times 10^{-111}$ & Yes \\
Common & Rarity Similarity (McNemar) & $0.4531$ & No \\ \hline
All Trials & Criteria Similarity ($t$-test) & $5.575 \times 10^{-145}$ & Yes \\
All Trials & Criteria Similarity (Wilcoxon) & $4.863 \times 10^{-109}$ & Yes \\
All Trials & Axis Similarity (McNemar) & $1.722 \times 10^{-200}$ & Yes \\
All Trials & Rarity Similarity (McNemar) & $2.944 \times 10^{-36}$ & Yes \\ \hline
\end{tabular}
\end{table*}

\end{document}